\documentclass{article} 
\usepackage{iclr2025_conference,times}

\usepackage{amsmath,amsfonts,bm}

\def\eqref#1{equation~\ref{#1}}

\def\1{\bm{1}}

\DeclareMathAlphabet{\mathsfit}{\encodingdefault}{\sfdefault}{m}{sl}
\SetMathAlphabet{\mathsfit}{bold}{\encodingdefault}{\sfdefault}{bx}{n}

\usepackage{hyperref}
\usepackage{url}

\usepackage{bm}

\usepackage{url}

\usepackage{multirow}
\usepackage{float} 
\usepackage[normalem]{ulem}
\usepackage{subcaption}
\usepackage{multirow}
\usepackage{booktabs}
\usepackage{tabularx}
\usepackage{multirow}
\usepackage{makecell}
\usepackage{framed,enumitem} 
\usepackage{amsmath,amssymb}

\usepackage{graphicx}
\usepackage{amssymb} 
\usepackage{bbding}
\usepackage{wrapfig}
\usepackage{adjustbox}
\usepackage{graphicx}
\usepackage{listings}
\usepackage{mathtools}
\usepackage{pdflscape}          

\usepackage{lipsum}

\usepackage{array}
\usepackage{rotating}

\newcolumntype{C}[1]{>{\centering\arraybackslash}m{#1}}
\newcolumntype{R}[1]{>{\raggedleft\arraybackslash}m{#1}}
\newcolumntype{L}[1]{>{\raggedright\arraybackslash}m{#1}}

\usepackage{soul}

\title{Decision Trees That Remember: Gradient-Based Learning of Recurrent Decision Trees with Memory}

\author{Sascha Marton\thanks{Equal Contribution} \\
University of Mannheim \\
\texttt{sascha.marton@uni-mannheim.de} \\
\And
Moritz Schneider$^{*}$ \\
Boehringer Ingelheim\\
\texttt{moritz.schneider@boehringer-ingelheim.com} \\
}

\iclrfinalcopy 
\begin{document}

\maketitle

\begin{abstract}
Neural architectures such as Recurrent Neural Networks (RNNs), Transformers, and State-Space Models have shown great success in handling sequential data by learning temporal dependencies. Decision Trees (DTs), on the other hand, remain a widely used class of models for structured tabular data but are typically not designed to capture sequential patterns directly. Instead, DT-based approaches for time-series data often rely on feature engineering, such as manually incorporating lag features, which can be suboptimal for capturing complex temporal dependencies.
To address this limitation, we introduce ReMeDe Trees, a novel recurrent DT architecture that integrates an internal memory mechanism, similar to RNNs, to learn long-term dependencies in sequential data. Our model learns hard, axis-aligned decision rules for both output generation and state updates, optimizing them efficiently via gradient descent. We provide a proof-of-concept study on synthetic benchmarks to demonstrate the effectiveness of our approach.
\end{abstract}

\begin{figure}[H]
    \centering
    \includegraphics[width=1\linewidth]{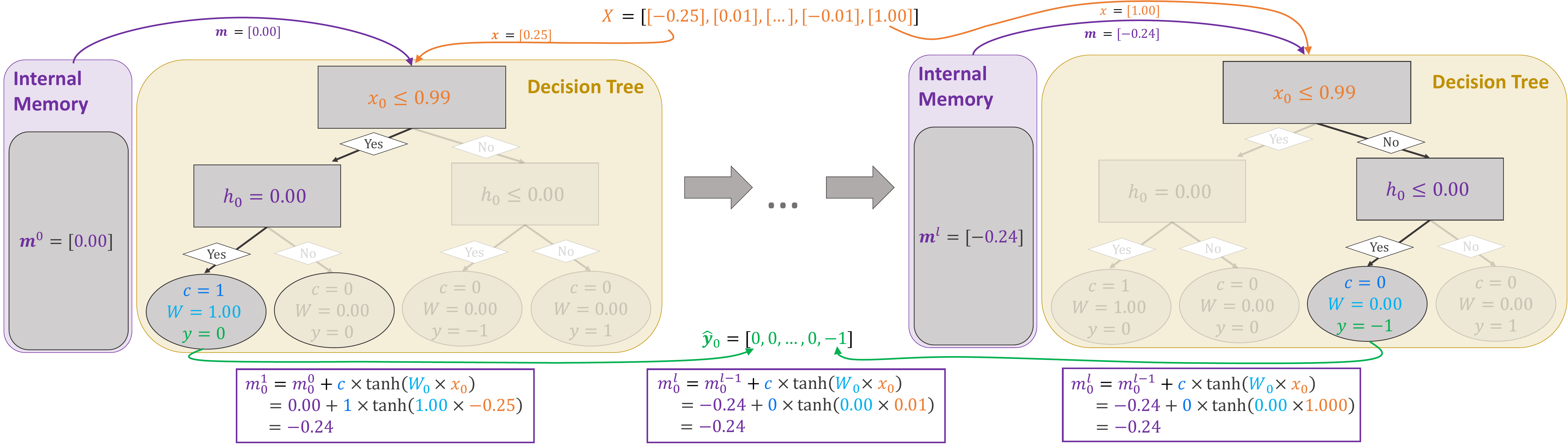}
    \caption{\textbf{Minimal Recurrent Decision Tree Example}
This figure shows an exemplary ReMeDe tree applied to a sign recognition task. The task is to memorize the sign of \( x \in (-0.5,0.5) \) at the first position and predict it (-1 or 1) when a trigger value (1) appears, while intermediate positions hold zeros plus small noise.
The figure depicts the minimal ReMeDe tree solving this task. At the root node, the tree checks whether the trigger occurs. If not (left branch), there are two cases: If the hidden state is zero, it updates based on input, adopting the sign of the entry; otherwise, it remains unchanged.
If the trigger occurs (right branch), the tree splits on the hidden state to predict the sign of the first value: negative for a negative hidden state, positive otherwise.
}
    \label{fig:remede_example}
\end{figure}

\section{Introduction}
Dealing with sequential, i.e. time-dependent, tabular data is an important area of machine learning research. Besides forecasting, dynamic modeling is crucial for data-driven control methods. Both have many practical applications in science, finance, healthcare, and many industrial areas. 

Generally speaking, there are two distinct structural approaches to learn dependencies over time. 
The often employed \emph{memory window} approach, that can be used with any type of regression or classification algorithm, reduces the temporal dependencies to a static prediction problem by collapsing the past $L$ (input or output) values within a time series into a flat input to the model. This is also sometimes referred to as (Nonlinear) Autoregressive Exogenous Model ((N)ARX)
\citep{nelles2020nonlinear}.
The other approach are \emph{recurrent} models, which deal with time dependency explicitly. Modern forms of recurrent architectures like Recurrent Neural Networks (RNNs)~\citep{elman1990finding}, Long Short-Term Memory networks (LSTMs)~\citep{schmidhuber1997long} 
define a hidden memory state, which is updated in each inference step together with the calculation of the model outputs. 

Recurrent approaches are in principle more powerful, because the model can deal with long-term dependencies exceeding any practical choice of $L$ for the memory window approach. Unfortunately, truly recurrent (neural) models are still challenging to train due to unstable dynamics of backpropagated gradients over long sequences \citep{hochreiter1998vanishing},
even with advances like gated memory units as in LSTM networks \citep{schmidhuber1997long}.
In addition, training neural networks can afford a large amount of data. In real-world applications with limited data availability, they are often outperformed by tree-based ensembles such as XGBoost \citep{chen2016xgboost} or CatBoost \citep{prokhorenkova2018catboost}. Unfortunately, for sequential data, such approaches have to be used with the limited memory window technique.

In this paper, we introduce a novel decision tree (DT) algorithm, \textbf{Re}current \textbf{Me}mory \textbf{De}cision (ReMeDe) Trees, that, for the first time, incorporates recurrence in DTs through an internal memory mechanism. Building on the techniques proposed by \citet{marton2024gradtree}, our method enables efficient training of DTs via gradient descent, resulting in hard, axis-aligned recurrent DTs capable of handling sequential data through a learnable internal memory. To the best of our knowledge, this is the first approach to learn a memory-augmented recurrent DT model using backpropagation through time \citep{werbos1990backpropagation}.
Specifically, our contributions are:
\begin{itemize}
    \item We extend Gradient-Based Decision Trees~\citep{marton2024gradtree} by incorporating an internal memory mechanism that can be learned using backpropagation through time.
    \item We modify the internal nodes of DTs to enable splits based on internal memory values, allowing pathing decisions to be conditioned on past experiences.
    \item We propose a novel update procedure for the internal memory, leveraging the DT's output at each time step and incorporating a hard memory gating mechanism.
\end{itemize}

First experiments with synthetic problems indicate that, similar to RNNs, ReMeDe Trees can overcome the limitations of fixed-size memory windows by efficiently compressing information in their hidden state. This suggests that ReMeDe Trees could offer a promising approach for time series tasks involving long-term dependencies, potentially combining the benefits of recurrent models for sequential data with the interpretability and axis-aligned structure of DTs. 

\section{Background}
In this section, we will introduce the foundational concepts for ReMeDe Trees, which includes the core notation and methodology of Gradient-Based Decision Trees, as well as Recurrent Neural Networks.

\subsection{GradTree: Gradient-Based Decision Trees}

This section introduces the core principles and notation of Gradient-Based Decision Trees (GradTree), which serve as the foundation for learning DTs through gradient-based optimization. For a comprehensive overview, we refer to \citet{marton2024gradtree}.

Traditional DTs rely on a hierarchical structure of nested decision rules. GradTree reformulates DTs into arithmetic functions based on addition and multiplication, enabling efficient gradient-based learning. Specifically, GradTree focuses on learning fully-grown (i.e., complete and balanced) DTs, which can later be pruned if necessary. This means that every node has either zero or two successors and all nodes with zero successors have the same depth. 

Such a tree of depth $d$ can be expressed in terms of its parameters as:
\begin{equation}\label{eq:tree}
\boldsymbol{y} = t(\boldsymbol{x} |  \boldsymbol{\lambda}, \boldsymbol{\tau}, \boldsymbol{\iota}) = \sum_{l=0}^{2^d-1} \lambda_l \, \mathbb{L}(\boldsymbol{x} | l, \boldsymbol{\tau}, \boldsymbol{\iota})
\end{equation}

Here, $\mathbb{L}$ is an indicator function that determines whether a data point \(\boldsymbol{x} \in \mathbb{R}^n\) reaches leaf node \(l\), \(\boldsymbol{\lambda} \in \mathcal{C}^{2^d}\) assigns class labels \(y \in Y\) to each leaf, \(\boldsymbol{\tau} \in \mathbb{R}^{2^d-1}\) contains the split thresholds, and \(\boldsymbol{\iota} \in \mathbb{N}^{2^d-1}\) specifies the feature index for each internal node. The output space  \(Y\)
may be a set of discrete class labels, in which $Y \subset \mathbb{N}^{n_y}$, or some continuous space \(Y \subset \mathbb{R}^{n_y}\) for application to regression problems. 

To enable gradient-based optimization and efficient computation using matrix operations, GradTree introduces a dense representation of DTs. The traditional feature index vector \(\boldsymbol{\iota}\) is expanded into a one-hot encoded matrix \(\boldsymbol{I} \in \mathbb{R}^{(2^d-1) \times n}\), and the split thresholds are represented as a matrix \(\boldsymbol{T} \in \mathbb{R}^{(2^d-1) \times n}\), allowing individual thresholds for each feature.
With internal nodes ordered in a breadth-first manner, the tree function can be reformulated as:
\begin{equation}\label{eq:gradtree}
g(\boldsymbol{x} | \boldsymbol{\lambda}, T, I) = \sum_{l=0}^{2^d-1} \lambda_l \, \mathbb{L}(\boldsymbol{x} | l, \boldsymbol{T}, \boldsymbol{I})
\end{equation}

The indicator function $\mathbb{L}$ for a leaf node \(l\) is defined as:
\begin{align}\label{eq:indicator_func}
\mathbb{L}(\boldsymbol{x} | l, \boldsymbol{T}, \boldsymbol{I}) = \prod^d_{j=1} \left(1 - \mathfrak{p}(l,j) \right) \mathbb{S}(\boldsymbol{x} | \boldsymbol{I}_{\mathfrak{i}(l,j)}, \boldsymbol{T}_{\mathfrak{i}(l,j)}) + \mathfrak{p}(l,j) \left(1 - \mathbb{S}(\boldsymbol{x} | \boldsymbol{I}_{\mathfrak{i}(l,j)}, \boldsymbol{T}_{\mathfrak{i}(l,j)}) \right)
\end{align}

In this formulation, $\mathfrak{i}(l,j)$ denotes the internal node on the path to leaf \(l\) at depth \(j\), and $\mathfrak{p}(l,j)$ indicates whether the path follows the left (\(\mathfrak{p} = 0\)) or right (\(\mathfrak{p} = 1\)) child node.

Traditional DTs use the non-differentiable Heaviside step function for splits, which impedes gradient-based learning. GradTree replaces this with a smooth approximation using the logistic sigmoid function:
\begin{equation}
     \mathbb{S}(\boldsymbol{x}| \boldsymbol{\iota}, \boldsymbol{\tau}) = \left\lfloor S \left( \boldsymbol{\iota} \cdot  \boldsymbol{x} - \boldsymbol{\iota} \cdot \boldsymbol{\tau} \right)  \right\rceil 
     \label{eq:split_sigmoid_round}
\end{equation}

where \(S(z) = \frac{1}{1 + e^{-z}}\) is the sigmoid function, \(\left\lfloor z \right\rceil\) rounds \(z\) to the nearest integer, and \(\boldsymbol{\iota} \cdot \boldsymbol{x}\) denotes the dot product. To maintain axis-aligned splits, \(\boldsymbol{\iota}\) is enforced as a one-hot encoded vector using a hardmax transformation.

Since both rounding and hardmax operations are non-differentiable, GradTree utilizes the straight-through (ST) estimator \citep{yin2019understandingstraightthroughestimatortraining} during backpropagation. This allows non-differentiable operations in the forward pass while enabling gradient flow in the backward pass.
In contrast to many approaches to learn soft, differentiable DTs, e.g. \citep{irsoy2012soft,luo2021sdtr}, GradTrees are structurally (and also w.r.t. the inference process) equivalent to classical DTs without necessity for any postprocessing, which may degrade the performance of the final DT model.

\subsection{Recurrent Neural Networks (RNNs)}
Recurrent Neural Networks (RNNs) are a class of neural networks designed to handle sequential data by maintaining a dynamic hidden state that captures temporal dependencies. 
Unlike feedforward neural networks, which assume independent input samples, RNNs incorporate recurrent connections, enabling them to retain information from previous time steps and model temporal correlations within sequences.

An RNN processes an input sequence \(\{\boldsymbol{x^1}, \boldsymbol{x^2}, \dots, \boldsymbol{x^t}\}\), where \(\boldsymbol{x^t} \in \mathbb{R}^{n_x}\) is the input vector at time step \(t\), by recursively updating a hidden state \(\boldsymbol{h^t} \in \mathbb{R}^{n_h}\) as follows:
\begin{equation}\label{eq:rnn_update}
    \boldsymbol{h^t} = \phi\left( W_{xh} \boldsymbol{x^t} + W_{hh} \boldsymbol{h^{t-1}} + b_h \right)
\end{equation}
where \(W_{xh} \in \mathbb{R}^{n_h \times n_x}\) is the input-to-hidden weight matrix,  \(W_{hh} \in \mathbb{R}^{n_h \times n_h}\) is the hidden-to-hidden weight matrix, \(b_h \in \mathbb{R}^{n_h}\) is the bias vector, and \(\phi(\cdot)\) is a activation function. Typically the hyperbolic tangent (\(\tanh\)) or ReLU are used, however, more recently using linear activation for the hidden state has also received increased attention \citep{orvieto2023resurrectingrecurrentneuralnetworks}. 
The network produces an output \(\boldsymbol{y}^t \in \mathbb{R}^{n_y}\) at each time step, computed as:
\begin{equation}\label{eq:rnn_output}
    \boldsymbol{y}^t = \psi\left( W_{hy} \boldsymbol{h}^t + b_y \right)
\end{equation}
where \(W_{hy} \in \mathbb{R}^{n_y \times n_h}\) is the hidden-to-output weight matrix, \(b_y \in \mathbb{R}^{n_y}\) is the output bias vector, and \(\psi(\cdot)\) is an activation function appropriate for the task, such as the softmax function for classification tasks.

RNNs are usually trained using Backpropagation Through Time (BPTT), an extension of the standard backpropagation algorithm that unfolds the network across time steps to compute gradients \citep{werbos1990backpropagation}, or Real-Time Recurrent Learning (RTRL) \citep{williams1989experimental}.
However, standard RNNs are prone to vanishing and exploding gradients, which limit their ability to learn long-term dependencies \citep{hochreiter1998vanishing}.
Despite these limitations, basic RNNs are effective for tasks involving short to moderate sequential dependencies and serve as a foundational model for more advanced recurrent architectures. Their parameter sharing across time steps allows efficient learning from sequences of varying lengths, making them applicable to time series prediction, text generation, and other sequential data modeling tasks.

Recognizing the limitations of basic RNNs, \citep{schmidhuber1997long} proposed an extension which makes use of a gating mechanism for the hidden state update. This involves additional gates for adding new and forgetting old information respectively, rendering the resulting model more capable of dealing with longer lag times.
This Long Short-Term Memory (LSTM) uses the following equations:
\begin{equation}
    \begin{aligned}
        f^t &= \sigma_g(W_f \boldsymbol{x^t} + U_f \boldsymbol{h^t} + b_f),\\
        i^t &= \sigma_g(W_i \boldsymbol{x^t} + U_i \boldsymbol{h^t} + b_i),\\
        o^t &= \sigma_g(W_o \boldsymbol{x^t} + U_o \boldsymbol{h^t} + b_o),\\
        \tilde{c}^t &= \sigma_c(W_c \boldsymbol{x^t} + U_c \boldsymbol{h^t} + b_c),\\
        c^t &= f^t \odot c^{t-1} + i^t \odot \tilde{c}^t,\\
        \boldsymbol{h^t} &= o^t \odot \sigma_h(c^t),
    \end{aligned}
\end{equation}
where $\boldsymbol{x^t} \in \mathbb{R}^{n_x}$ denotes the input vector at time step $t$, $\odot$ denotes the Hadamard (elementwise) product,
$i^t$ is the input gate activation, $o^t$ is the output gate activation, $\boldsymbol{h^t}$ is the hidden state vector, $\tilde{c^t}$ is the cell state activation, $c^t$ is the cell state, $W_{f,i,o,c},U_{f,i,o,c}$ are weight matrices, and $b_{f,i,o,c}$ are bias vectors.
Furthermore, $\sigma_g$ is a sigmoid function, and $\sigma_{c,h}$ is \(\tanh\).

\section{Recurrent Memory Decision Trees}
As in the previous section, we consider time-series problems, where for each time step $k=1,2,...$, a value $\boldsymbol{x^k} \in \mathbb{R}^{n_x}$ is observed.
The outputs $\boldsymbol{y^k} \in Y$ may be either continuous, where $Y \subset \mathbb{R}^{n_y}$, or discrete in which $Y \subset \mathbb{N}^{n_y}$.

\paragraph{The Hidden Memory}
There are two approaches to deal with time-depencency in dynamic models: 
(1) NARX models where information about $L$ past inputs $\boldsymbol{x^{k-1}},...,\boldsymbol{x^{k-L}}$ or outputs $\boldsymbol{y^{k-1}},...,\boldsymbol{y^{k-L}}$ is used as model input in time step $k$.
(2) Recurrent models where information about an indefinite number of past time steps is used as model input by means of an additional $n_m$-dimensional memory $\mathrm{M} \subset \mathbb{R}^{n_m}$, which stores (compressed) information about the past. Here, $\boldsymbol{\mathrm{m}^k}$ is treated as an input variable to determine $\boldsymbol{y^k}$, but it also will be updated by the model in each inference step.

It is obvious that NARX models can only effectively model Markov Processes up to order $L$, whereas recurrent models may be able to deal with much higher information lags.
In order to use a hidden memory $\mathrm{M}$ in a DT, it has to be observed by internal nodes, similar to $X$, and modified by leaf nodes, similar to $Y$.
This means that we can use the same equations as in GradTrees, but with $\tilde{X}=X \times \mathrm{M}$ and $\tilde{Y}=Y \times \mathrm{M}$.
Since $\boldsymbol{\tilde{y}}=(\boldsymbol{y},\boldsymbol{m})$, we may write
\begin{equation}\label{eq:remede_output}
\begin{aligned}
    \boldsymbol{y^{t}}=g(\boldsymbol{\tilde{x}^t} | \boldsymbol{\lambda}, T, I)_y,\\
    \boldsymbol{m^{t}}=g(\boldsymbol{\tilde{x}^t} | \boldsymbol{\lambda}, T, I)_m.
\end{aligned}
\end{equation}

Assuming that the memory is initially set to all zeros, i.e. $\boldsymbol{\mathrm{m}^0} = \boldsymbol{0}_{n_m}$, we retain the general structure and methodology as in GradTree that
can be trained by gradient descent (or, to be more precise, in this case: Backpropagation-Through-Time, see \citet{werbos1990backpropagation}).


\paragraph{Internal Decision Nodes}
Similar to classical DTs and GradTree, we currently employ hard, axis-aligned splits. 
However, ReMeDe Trees operate in the combined input-state space $\tilde{X} = X \times M$.
This means that at each internal decision node, a routing decision through the tree may be either based on a component of the input vector, or a particular dimension of the hidden memory state. Hence the inference logic within the tree may explicitly depend on stored past information.

\paragraph{Memory Gating}
Gating techniques have been introduced in RNNs to deal with unstable gradient dynamics during training \citep{hochreiter1998vanishing}.
Therein, additional input- or state-dependent gates determine write-operations to the hidden state (either update with new information, or even deletion of old information \citep{schmidhuber1997long}), as formalized in the previous section. 
Augmenting the memory access operation in ReMeDe Trees with gating mechanisms is quite straightforward and should - similar to their effect in RNNs - allow the model to deal better with longer dependencies over time. 
We use a very simple form of non-smooth, i.e. binary gating which aligns very well with the overall DT model structure, and leave studying more intricate mechanisms for future work. This gating mechanism will be introduced along with the output representation in the next paragraph. 

\paragraph{Output Representation}
For ReMeDe Trees, each leaf node prescribes an output value, but also an update to the ${n_m}-$dimensional memory state $\boldsymbol{m^t}$.
Classical DTs use a zero-order output representation, i.e. the output value is explicitly prescribed in the leaf nodes. For classification tasks, such as those considered in the experiments within this article, this is of course reasonable. 
Hence, the output will be calculated as
\begin{equation}\label{eq:remede_out}
    \boldsymbol{y^{t}}=g(\boldsymbol{\tilde{x}^t} | \boldsymbol{\lambda}, T, I)_y = y_j,
\end{equation}
where $\boldsymbol{y_j} \in Y$ denotes the constant output prescribed in the leaf node $j$ that was selected by the tree inference.
However, for continuous output values - such as the memory updates - other variants have been considered. For applications in time-series prediction, it is often recommended in practice to use a first order output, i.e.
\begin{equation}\label{eq:remede_fo_output}
    \boldsymbol{y^{t}}=\boldsymbol{y^{t-1}} + g(\boldsymbol{\tilde{x}^t} | \boldsymbol{\lambda}, T, I)_y,
\end{equation}
to be able to deal with trends effectively.
Other, more sophisticated, approaches utilize a parametrized mapping in each leaf node for static or dynamic problems, such as linear model trees \citep{czajkowski2016role,ammari2023linear} or fuzzy weighted linear models in the LoLiMoT algorithm \citep{nelles1996basis}.
We consider studying different variants of output representation, in particular for memory updates, an interesting avenue for future research. For outputs, we use a zero order formulation and for the memory update an RNN-inspired parametrized equation:
\begin{equation}
\begin{aligned}
   \boldsymbol{m^{t}} &= \boldsymbol{m^{t-1}} + \left\lfloor \psi_g(c_j) \right\rceil \psi( W^x_j \boldsymbol{x^t}),\\
   c_j,W^x_j &= g(\boldsymbol{\tilde{x}^t} | \boldsymbol{\lambda}, T, I)_m,
\end{aligned}
\end{equation}
where $j$ denotes the leaf node selected by tree inference, $\psi$ is \(\tanh\), $W^x_j \in \mathbb{R}^{n_m \times n_x}$ is a learnable weight matrix, $c_j$ is a zero order output prescribed by leaf node $j$, ${\psi_g: \mathbb{R} \rightarrow [0,1]}$ is a sigmoid function, and ${\left\lfloor \cdot \right\rceil: \mathbb{R} \rightarrow \mathbb{Z}}$ maps its argument to the nearest integer, i.e. applied componentwise to $\psi_g(c_j)$, we have ${\left\lfloor \psi_g(c_j) \right\rceil \in \{0,1\}^{n_m}}$, representing the hard gating mechanism for the hidden state update. Similar to the split decision in GradTree, this is achieved by rounding the sigmoid output of a gating parameter $c_j$ and using the ST operator to ensure a reasonable gradient flow.

\section{Evaluation}
In our evaluation, we provide a proof of concept that our formulation allows learning a recurrent memory DT architecture in an RNN-like fashion solely with BPTT.
Therefore, we evaluate our method on 5 different synthetic datasets with increasing complexity that are designed in a way that they can only be solved with an internal memory,
whenever a competing memory window-based model faced hard limits with respect to window size.

\subsection{Proof of Concept - Synthetic Data Generation Procedures} \label{sec:poc}
This subsection introduces five synthetic data generation procedures designed to model temporal dependencies and delayed response behaviors in time series data. Each method simulates distinct patterns, including delayed reactions and memory effects, across one- and two-dimensional input spaces. The following subsections describe each method in detail with corresponding mathematical formalizations.

\paragraph{1. Delayed Sign Retrieval (Single-Dimensional, Fixed Delay)}

The first procedure generates a single-dimensional time series where the task is to recover the sign of the initial input after a fixed delay. A trigger signal appears at a specific timestep, prompting the output to reflect the sign of the initial value, while the output remains zero at all other timesteps.
Let \( x_0 \sim \mathcal{U}(-v, v) \) be the initial value, and \( d \) denote the fixed delay. The input sequence \( \mathbf{x} \in \mathbb{R}^{d+2} \) and the target output \( \mathbf{y} \in \mathbb{R}^{d+2} \) are defined as:
\begin{equation}
\mathbf{x} = [x_0, 0, 0, \dots, 0, t],
\end{equation}
\begin{equation}
\mathbf{y} = [0, 0, \dots, 0, \operatorname{sign}(x_0)],
\end{equation}

where \( t \) is the trigger value and the sign function is defined as:

\begin{equation}
\operatorname{sign}(x) = \begin{cases}
1 & \text{if } x \geq 0, \\
-1 & \text{if } x < 0.
\end{cases}
\end{equation}

\paragraph{2. Delayed Sign Retrieval (Two-Dimensional, Fixed Delay)}

This method extends the previous setup to a two-dimensional input. The first channel contains the initial value, and the second channel receives the trigger after a fixed delay. The model must output the sign of the first input upon the appearance of the trigger.
Let \( x_0 \sim \mathcal{U}(-v, v) \) and \( d \) be the fixed delay. The input matrix \( \mathbf{X} \in \mathbb{R}^{(d+2) \times 2} \) and the output \( \mathbf{y} \in \mathbb{R}^{d+2} \) are defined as:
\begin{equation}
\mathbf{X} = \begin{bmatrix}
x_0 & 0 \\
0 & 0 \\
\vdots & \vdots \\
0 & t
\end{bmatrix}, \quad
\mathbf{y} = [0, 0, \dots, 0, \operatorname{sign}(x_0)].
\end{equation}

\paragraph{3. Delayed Sign Retrieval (Single-Dimensional, Variable Delay)}

This variant introduces a variable delay, randomly sampled from a uniform range \([d_{\text{min}}, d_{\text{max}}]\). The trigger appears at a random timestep, requiring the model to output the sign of the initial value.
Let \( \delta \sim \mathcal{U}(d_{\text{min}}, d_{\text{max}}) \) and \( x_0 \sim \mathcal{U}(-v, v) \). The input \( \mathbf{x} \) and output \( \mathbf{y} \) are defined as:

\begin{equation}
\mathbf{x} = [x_0, 0, \dots, 0, t, 0, \dots],
\end{equation}
\begin{equation}
\mathbf{y} = [0, \dots, 0, \operatorname{sign}(x_0), 0, \dots],
\end{equation}

where the trigger \( t \) appears at timestep \( \delta \).

\paragraph{4. Delayed Sign Retrieval (Two-Dimensional, Variable Delay)}

This method generalizes the two-dimensional fixed delay scenario by allowing the trigger to appear at a randomly chosen timestep within a predefined delay range.
Let \( \delta \sim \mathcal{U}(d_{\text{min}}, d_{\text{max}}) \). The input matrix \( \mathbf{X} \) and output \( \mathbf{y} \) are:
\begin{equation}
\mathbf{X} = \begin{bmatrix}
x_0 & 0 \\
0 & 0 \\
\vdots & \vdots \\
0 & t \\
\vdots & 0
\end{bmatrix}, \quad
\mathbf{y} = [0, \dots, 0, \operatorname{sign}(x_0), 0, \dots].
\end{equation}

\paragraph{5. Sign Memory Task}

The final procedure generates sequences composed of alternating blocks of \( -1 \) and \( 1 \), interspersed with zero-valued delay blocks. The task is to reproduce the last non-zero block upon encountering a new non-zero block.
Let \( b_j \in \{-1, 1\} \) denote the \( j \)-th non-zero block of length \( l \), and \( z \) represent a zero block of length \( d \). The input \( \mathbf{x} \) and target output \( \mathbf{y} \) are constructed as:
\begin{equation}
\mathbf{x} = [b_1, z, b_2, z, \dots, b_n],
\end{equation}
\begin{equation}
\mathbf{y} = [0, z, b_1, z, b_2, \dots].
\end{equation}

Each non-zero block \( b_j \) is randomly selected from \(\{-1, 1\}\), and the model must recall and reproduce the previous block at the appropriate timestep.

\subsection{Experimental Setup}

\paragraph{Methods} We evaluate two recurrent architecture on our datasets: LSTMs and ReMeDe Trees. While more recent RNNs, such as xLSTM \citep{beck2024xlstm}, might provide a SoTA-benchmark, the aim here is to show viability of the approach instead of benchmarking by comparing two recurrent models with gated hidden state updates. 
We omit comparison with NARX models, as it is clear that given a fixed lookback size of the model, our experiments can always be configured such that these models cannot learn the necessary temporal dependencies.
Furthermore, we compare against two baselines, a simple random guess and a naive baseline making an informed guess (i.e., predicting the most probable value for each element in the sequence) based on the task.

\paragraph{Hyperparameters} To select suitable hyperparameters for each task, we used Optuna~\citep{akiba2019optuna} with 60 trials. Specifically, we optimized only the learning rates, while keeping all other hyperparameters fixed. In particular, we selected a small tree depth of $6$ and a hidden state size of only $5$, to demonstrate that even with a compact model architecture, meaningful patterns can still be learned effectively. For LSTM, we selected a basic architecture with two hidden layers of 32 and 16 neurons and dropout. Similar to ReMeDe, we optimized the learning rate using Optuna with 60 trials.

\paragraph{Datasets} For our proof-of-concept experiments, we utilized the datasets introduced in Section~\ref{sec:poc}. Specifically, we set the fixed delay to 5 and defined the variable delay within the range \([3,7]\). For each task, we generated a total of 10,000 sequences. Continuous values were sampled from the uniform distribution \(\mathcal{U}(-0.5, 0.5)\), and the delay was perturbed with a small random noise drawn from the normal distribution \(\mathcal{N}(-0.01, 0.01)\).

\begin{table*}[tb]
    \caption{\textbf{PoC Results.} We report the average test accuracy along with the standard deviation on our proof-of-concept datasets, computed over five independent random trials.}
    \label{tab:results_poc}
    \centering
    \resizebox{0.99\linewidth}{!}{
    \begin{tabular}{lrrrrr}
   
        \toprule
         & PoC 1 & PoC 2 & PoC 3 & PoC 4 & PoC 5 \\ \midrule
        ReMeDe (ours)              & \textbf{1.000 $\pm$ 0.000} & \textbf{1.000 $\pm$ 0.000} & \textbf{1.000 $\pm$ 0.000} & \textbf{1.000 $\pm$ 0.000} & \textbf{1.000 $\pm$ 0.000} \\
        LSTM              & \textbf{1.000 $\pm$ 0.000} & \textbf{1.000 $\pm$ 0.000} & \textbf{1.000 $\pm$ 0.000} & \textbf{1.000 $\pm$ 0.000} & \textbf{1.000 $\pm$ 0.000} \\
        Random Guess        & 0.334 $\pm$ 0.001 & 0.333 $\pm$ 0.001 & 0.332 $\pm$ 0.002 & 0.333 $\pm$ 0.001 & 0.333 $\pm$ 0.001 \\
        Na\"ive Baseline    & 0.930 $\pm$ 0.002 & 0.930 $\pm$ 0.002 & 0.889 $\pm$ 0.001 & 0.889 $\pm$ 0.002 & 0.877 $\pm$ 0.003 \\
        \bottomrule
    \end{tabular}
    }
\end{table*}

\subsection{Results}

\paragraph{We can learn recurrent decision trees with backpropagation through time}
Our results confirm that recurrent DTs can be effectively trained end-to-end using backpropagation through time. As shown in Table~\ref{tab:results_poc}, our ReMeDe model achieves perfect test accuracy across all PoC datasets, matching the performance of LSTM baselines. This demonstrates that gradient-based optimization using the proposed method is a viable approach for learning DTs with temporal dependencies, enabling both structured decision-making and sequence modeling within a single method.

\newpage

\begingroup
\setlength{\intextsep}{12pt}%
\begin{wraptable}{r}{0.415\textwidth}
\begin{minipage}{0.415\textwidth}
    \centering
    \caption{\textbf{Average Tree Size.} We report the average tree size, measured in terms of the number of nodes.} 
    \label{tab:tree_size}    
    \begin{tabular}{cc}
    
    \toprule
         & Number of Nodes \\ \midrule
        PoC 1 & 22.2 \\
        PoC 2 & 20.2 \\
        PoC 3 & 21.0 \\
        PoC 4 & 23.0 \\
        PoC 5 & 43.8 \\ \midrule
        Mean & 26.0 \\
        \bottomrule
    \end{tabular}
\end{minipage}
\end{wraptable}
\paragraph{ReMeDe Trees have a small tree size}
Table~\ref{tab:tree_size} presents the average tree size, measured in terms of the number of nodes, including both internal and leaf nodes, across our proof-of-concept datasets. The DTs are pruned by removing all redundant paths, ensuring a more compact representation. The results indicate that the learned DTs remain compact, with an average size of 26.0 nodes. Notably, the tree learned on PoC5 exhibits a considerably larger size, averaging 43.8 nodes, whereas the trees for the remaining tasks are of similar size, ranging between 20 and 23 nodes. This observation underscores the efficiency of the proposed method in capturing underlying dependencies while maintaining a moderate tree size. The compactness of ReMeDe Trees is particularly advantageous for interpretability on small datasets and enhances verifiability for more complex tasks.

\begin{figure*}[t]
    \centering
    \includegraphics[width=1\linewidth]{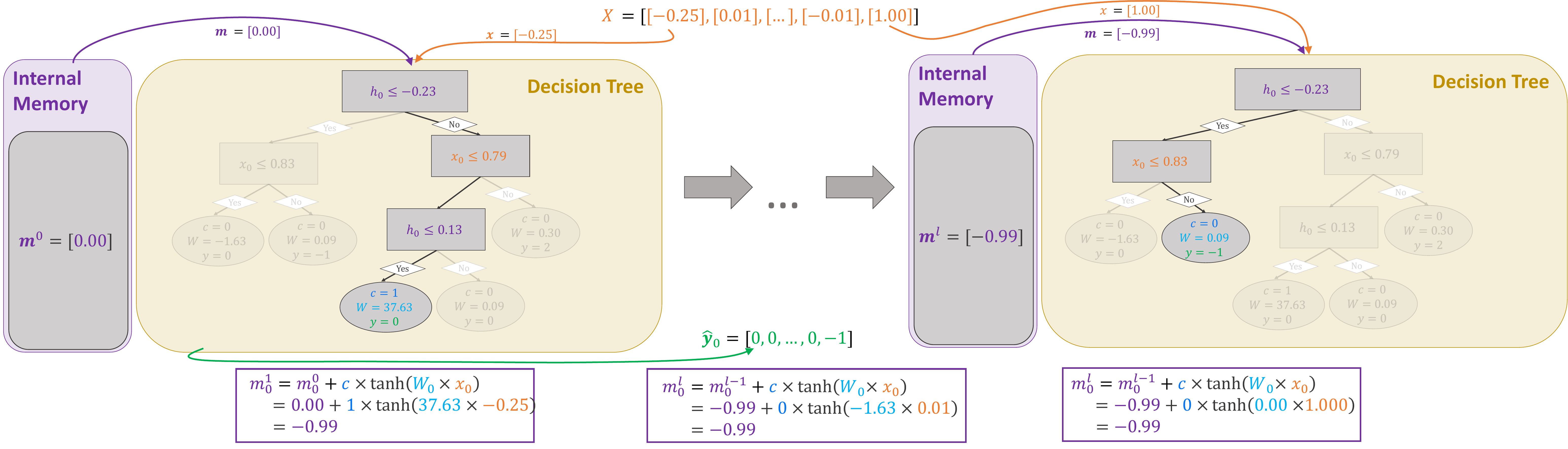}
    \caption{\textbf{ReMeDe Tree Update Visualization}
This figure shows an ReMeDe tree trained to a sign recognition task. The task is to memorize the sign of \( x \in (-0.5,0.5) \) at the first position and predict it (-1 or 1) when a trigger value (1) appears, while intermediate positions hold zeros plus small noise.}
    \label{fig:remede_runthrough}
\end{figure*}

\paragraph{ReMeDe Trees can effectively update and access the internal memory}
To illustrate how state updates operate in a compact ReMeDe tree, we present the example in Figure~\ref{fig:remede_runthrough}. The tree depicted in this figure was learned by our method in a simplified setting (using a tree of depth 4 with a single memory parameter) on PoC1. 
At the root node, the tree evaluates whether the hidden state is smaller than \(-0.23\), effectively distinguishing whether the first entry in the sequence was negative (left branch) or positive (right branch). At the second level, the tree checks whether the trigger condition is met (\(> 0.5\)). If the trigger is activated, the tree makes the corresponding prediction for the sign. Otherwise, the hidden state may be updated.  
The update mechanism follows the left path in the diagram, where the hidden state is updated only if it falls within the interval \([-0.23, 0.13]\). This condition is typically satisfied only for the first element in the sequence, as subsequent updates are significantly amplified by a weight of \(37.63\). If the hidden state lies outside this interval, no update occurs, which corresponds to the delay phase.
This example highlights how ReMeDe effectively captures and recalls sequential information, demonstrating its suitability for structured decision-making in temporal tasks.





\section{Related Work}
Classical DT learning algorithms, such as C4.5 \citep{quinlan2014c4} or CART \citep{breiman2017classification}, are based on growing a DT by greedily splitting the input space
in a componentwise fashion to optimize the reduction in the chosen error metric at each step of building the tree.
No method has been yet proposed to incorporate updates of an internal memory state based on these algorithms.

Nevertheless, the idea of using explicit time-dependency in the DT framework is not new.
\citep{chen2016learning} propose a model which they call recurrent DT, for camera planning. In contrast to ReMeDe, no internal memory state is used but previous outputs are fed back into the model as inputs, which renders this approach a special case of NARX models in the terminology used here. 
The same holds for \citep{chegini2010prediction}, who extend the LoLiMoT algorithm \citep{nelles1996basis} to include output feedback for financial time series prediction.
\citep{alaniz2021learningdecisiontreesrecurrently} propose an intricate scheme to learn a recurrent model, involving a DT, but also a combination between an LSTM and an Attribute-Learning System, where a DT uses the hidden state of an LSTM. 

Others have taken the converse route and combine classical DT with recurrent models in leaf nodes, such as \citep{ren2021tree}. Therein, first the input data is split using classical DT algorithms and then separate RNNs are trained for each leaf node, inheriting the potential suboptimality of the former.
Also worth mentioning is a family of approaches that uses hierarchical, tree structured switching linear systems for dynamics modeling, such as \citep{nassar2018tree}, which share some structural similarities with ReMeDe Trees, although the resulting models are quite different. In particular, the hidden state used there is discrete and some of the involved operations are soft, i.e. stochastic.
In contrast, a ReMeDe Tree consists only of a single hard, axis-aligned DT which performs read and write operations on its own hidden memory state, enabled by training the complete model via gradient descent. To the best of our knowledge, no other recurrent DT using continuous hidden state feedback has been proposed yet.

\section{Conclusion and Future Work}
In this article, we introduce a novel recurrent method, Recurrent Memory Decision (ReMeDe) Trees, which leverages an internal hidden state trained through Backpropagation-Through-Time to construct hard, axis-aligned and recurrent DTs building on the GradTree model \citep{marton2024gradtree}.
We have shown on synthetic test problems that our method is able to effectively compress past information into its hidden state to capture dependencies between inputs and outputs.

In the future, we would like to extend our method to more advanced base models, such as DTs with non-trivial output representations in leaf nodes and advanced memory gating techniques. Additionally, ReMeDe Trees can be readily introduced into tree ensembling approaches, such as GRANDE \citep{marton2024grande}.
Combining the basic ReMeDe Tree model presented in this paper with the aforementioned extensions may hopefully show that recurrent DTs have the potential to yield competitive performance in time series learning tasks involving long-term dependencies, combining the advantages of recurrent models in time series tasks with the advantages of hard, axis-aligned DTs.

\bibliography{iclr2025_conference}
\bibliographystyle{iclr2025_conference}

\end{document}